%% file: main.tex
\documentclass[letterpaper,10pt,conference]{ieeeconf}
\IEEEoverridecommandlockouts
\makeatletter
\let\NAT@parse\undefined
\makeatother

\usepackage{amssymb,amsmath}
\usepackage{amsthm}
\let\labelindent\relax
\usepackage{enumitem}
\usepackage{booktabs}
\usepackage[font=footnotesize,labelformat=simple]{subcaption}
\usepackage[font=footnotesize]{caption}

\usepackage{tikz}
\usepackage[mode=buildnew]{standalone}
\usepackage{xspace}
\usepackage{algorithmic}
\usepackage{algorithm}
\usepackage{tabulary}
\usepackage{verbatim}
\usepackage[numbers,sort&compress]{natbib}
\usepackage{thmtools}
\usepackage{mathtools}
\usepackage{multirow}
\usepackage{cuted}
\usepackage{hyperref}
\usepackage{multicol}
\usepackage{graphicx}
\hypersetup{
  pdfinfo={
    Title={Data Efficient Behavior Cloning for Fine Manipulation via Continuity-based Corrective Labels},
    Author={Abhay Deshpande, Liyiming Ke, Quinn Pfeifer, Abhishek Gupta, and Siddhartha S. Srinivasa},
    Subject={Imitation Learning},
    Keywords={Imitation Learning; Corrective Labels}
  },
  bookmarks=true,
  bookmarksopen=true,
  bookmarksnumbered=true,
  pdfpagemode=UseOutlines,
  colorlinks=true,
  linkcolor=blue,
  anchorcolor=blue,
  citecolor=blue,
  filecolor=blue,
  menucolor=blue,
  urlcolor=blue
}

\usepackage{lipsum}% just for the example

\makeatletter
\newcommand\fs@betterruled{%
  \def\@fs@cfont{\bfseries}\let\@fs@capt\floatc@ruled
  \def\@fs@pre{\vspace*{5pt}\hrule height.8pt depth0pt \kern2pt}%
  \def\@fs@post{\kern2pt\hrule\relax}%
  \def\@fs@mid{\kern2pt\hrule\kern2pt}%
  \let\@fs@iftopcapt\iftrue}
\floatstyle{betterruled}
\restylefloat{algorithm}
\makeatother

\theoremstyle{plain}
\newtheorem{theorem}{Theorem}[section]

\theoremstyle{definition}
\newtheorem{definition}[theorem]{Definition}

\theoremstyle{remark}
\newtheorem{remark}[theorem]{Remark}

\newcounter{hypnum}
\DeclareRobustCommand{\hypothesis}[1]{%
   \refstepcounter{hypnum}%
   \thehypnum\label{#1}}
\newcommand{\hypref}[1]{\textbf{H.\ref{#1}}}

\definecolor{Highlight}{cmyk}{1,0.19,0,.22}

\title{\LARGE \bf
Data Efficient Behavior Cloning for Fine Manipulation\\via Continuity-based Corrective Labels 
}

\author{
  Abhay Deshpande, Liyiming Ke, Quinn Pfeifer, Abhishek Gupta, Siddhartha S. Srinivasa%
  \thanks{
    Paul G. Allen School of Computer Science \& Engineering,
    University of Washington
    \texttt{\{abhayd, kayke, qpfeifer, abhgupta, siddh\}@cs.washington.edu}}%
}

\newcommand\blfootnote[1]{%
  \begingroup
  \renewcommand\thefootnote{}\footnote{#1}%
  \addtocounter{footnote}{-1}%
  \endgroup
}

\begin{document}
\maketitle
\thispagestyle{empty}
\pagestyle{empty}

\blfootnote{See more: \href{https://personalrobotics.github.io/CCIL/}{\textcolor{blue}{\texttt{https://personalrobotics.github.io/CCIL/}}}}

\input{inputs/0-abstract}

\input{inputs/1-introduction}
\input{inputs/3-method}
\input{inputs/4-experiment}
\input{inputs/5-results}
\input{inputs/6-discussion}
\input{inputs/2-related-work}
\input{inputs/999-acknowledgement}

{
\footnotesize
\renewcommand{\refname}{REFERENCES}
\bibliographystyle{IEEEtran}
\bibliography{main}
}

\end{document}

%% file: inputs/0-abstract.tex
\iffalse
\begin{strip}
\begin{minipage}{\textwidth}\centering
\vspace{-40pt}
\includegraphics[width=0.3\textwidth,height=0.4\textwidth]{media/cube.JPEG}
\includegraphics[width=0.3\textwidth,height=0.4\textwidth]{media/lego.JPEG}
\includegraphics[width=0.3\textwidth,height=0.4\textwidth]{media/coin.JPEG}
\captionof{figure}{Three fine manipulation tasks. From left to right: picking up a small cube, inserting a Lego gear into a mechanism, and picking up a coin.}
\label{fig:cover}
\end{minipage}
\end{strip}
\fi

\begin{strip}
\begin{minipage}{\textwidth}\centering
    \vspace{-40pt}
    \includegraphics[height=0.36\textwidth]{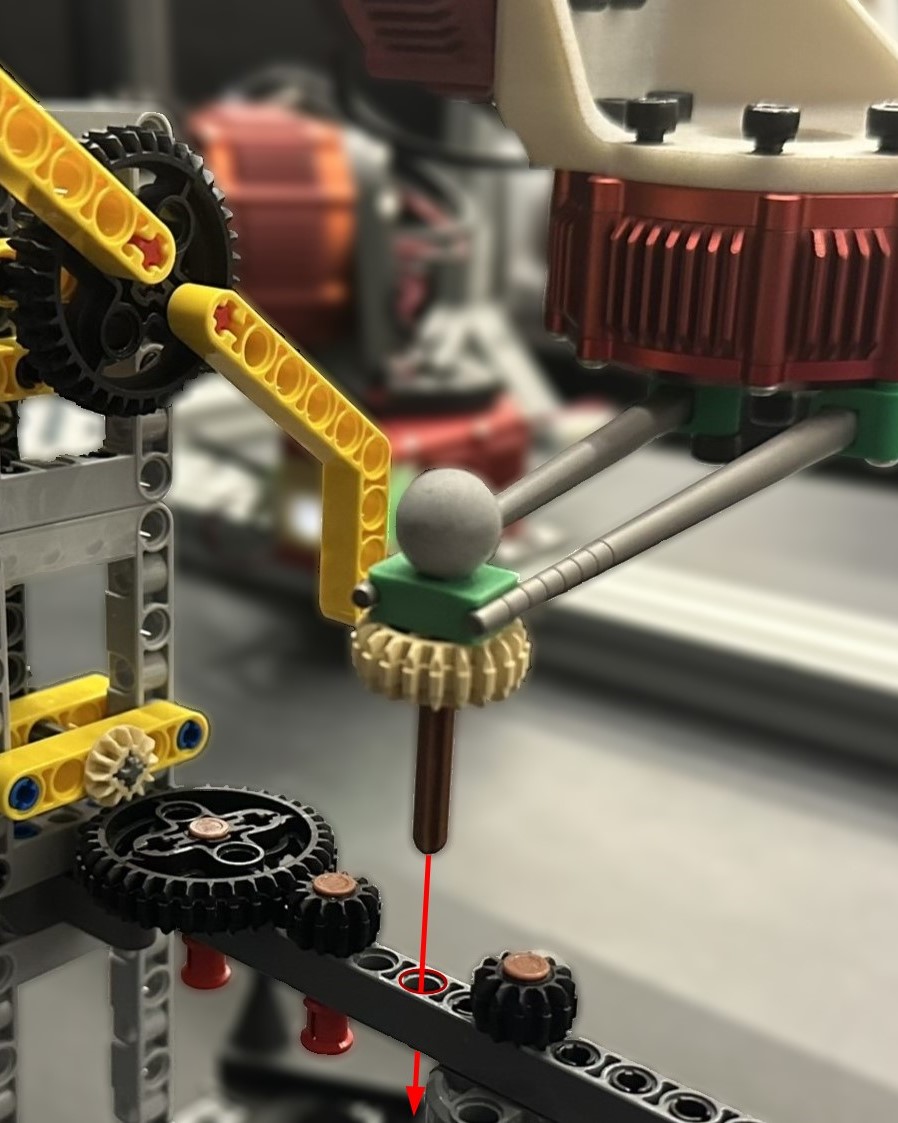}
    \includegraphics[height=0.36\textwidth]{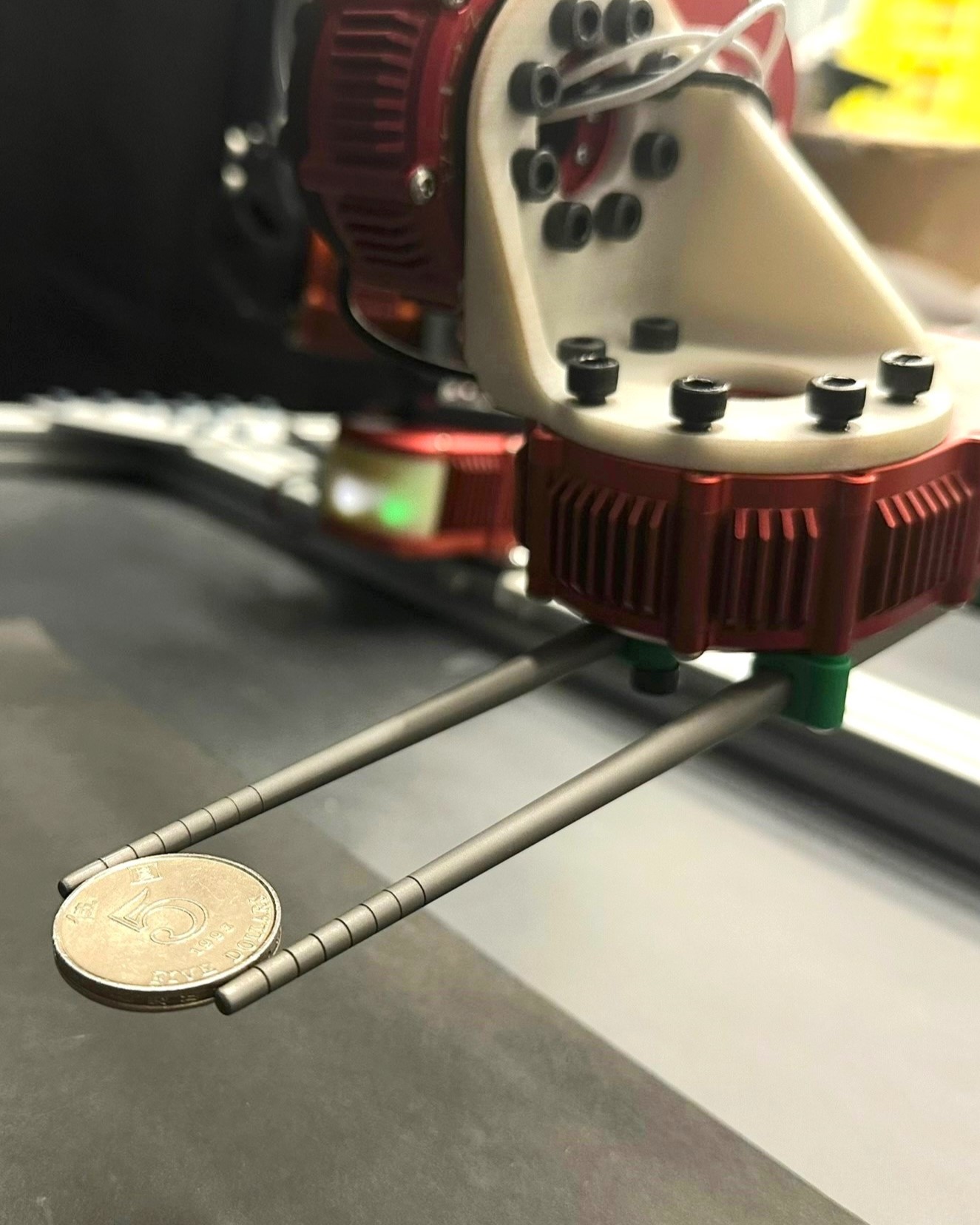}
    \includegraphics[height=0.36\textwidth]{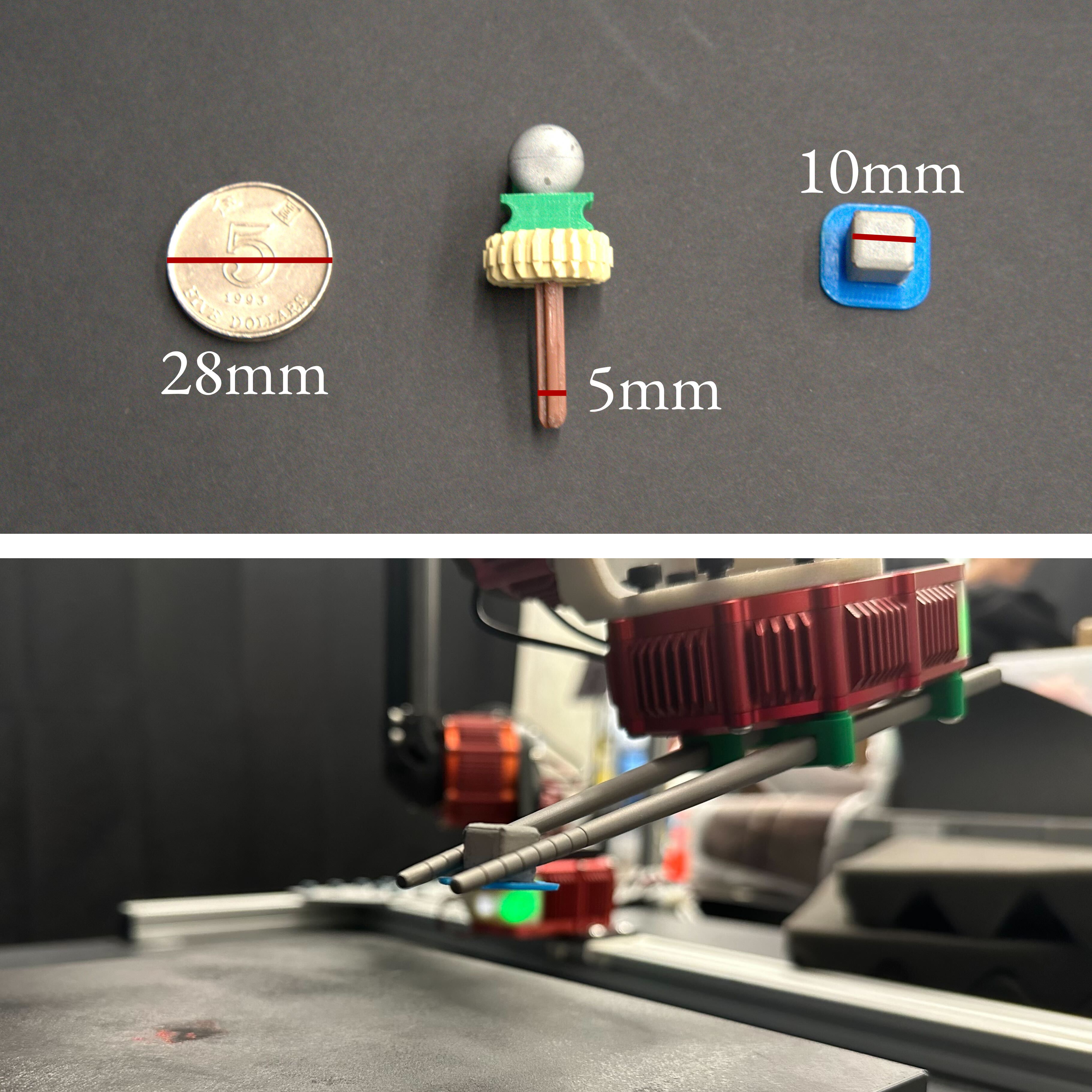}
        % \includegraphics[width=0.9\linewidth]{media/cover.jpg}
        % \caption{Our hardware platform and the various objects manipulated in our experiments, explained in detail in Fig.~\ref{fig:system}.}
        % \vspace{-.5em}
    \captionof{figure}{The three tasks we consider, \emph{GearInsertion}, \emph{GraspCoin}, and \emph{GraspCube}, along with our three task objects: a coin, a Lego gear, and a small cube.}
    \label{fig:front}
    \label{fig:task_desc}
    \label{fig:cover}
    \vspace{-.5em}
\end{minipage}
\end{strip}
%\sandy{Excellent Abstract!}
\begin{abstract}
We consider imitation learning with access only to expert demonstrations, whose real-world application is often limited by covariate shift due to compounding errors during execution. We investigate the effectiveness of the Continuity-based Corrective Labels for Imitation Learning (CCIL) framework in mitigating this issue for real-world fine manipulation tasks. CCIL generates corrective labels by learning a locally continuous dynamics model from demonstrations to guide the agent back toward expert states. Through extensive experiments on insertion and fine grasping tasks, we provide the first empirical validation that CCIL can significantly improve imitation learning performance despite discontinuities present in contact-rich manipulation. We find that: (1) real-world manipulation exhibits sufficient local smoothness to apply CCIL, (2) generated corrective labels are most beneficial in low-data regimes, and (3) label filtering based on estimated dynamics model error enables performance gains. To effectively apply CCIL to robotic domains, we offer a practical instantiation of the framework and insights into design choices and hyperparameter selection. Our work demonstrates CCIL's practicality for alleviating compounding errors in imitation learning on physical robots.
\end{abstract}

%% file: inputs/1-introduction.tex
\section{Introduction}

Imitation learning has shown remarkable scalability with large datasets~\citep{alvinn, floridi2020gpt}, emerging as a promising paradigm for robots to mimic complex behaviors from demonstrations~\citep{padalkar2023open, chi2023diffusion, billard2012imitation}. However, usage of imitation learning in robotics is often constrained by covariate shift and the scarcity of demonstrations~\citep{osa2018algorithmic, spencer2021feedback}. Real-world robotic policies can suffer from an accumulation of sensor and joint noises when a robot executes a trajectory. These compounding errors can cause the agent to encounter out-of-distribution states, resulting in task failures.  

Though various methods have been proposed to address this issue, their application can be limited by requirements such as the availability of interactive experts~\citep{ross2011reduction, bryan2019data}, simulators~\citep{wang2024cyberdemo}, a large-batch of sub-optimal data~\citep{chang2021mitigating}, and task-specific domain knowledge~\citep{venkatraman2014data, ke2021grasping, zhou2023nerf, chen2023genaug, block2024provable}. To achieve broad applicability without complex prerequisites, we focus on imitation learning that operates under practical assumptions, relying solely on offline demonstrations without requiring interactive experts or simulators. 

In this context, the framework of Continuity-based Corrective labels for Imitation Learning (CCIL)~\citep{ke2023ccil} has demonstrated promising results in simulations across multiple domains. CCIL combats compounding errors in imitation learning by generating corrective labels that bring the robot back to expert states. To do so without the ground truth dynamics model, CCIL learns a dynamics model from the demonstrations and uses it to synthesize labels. Importantly, it leverages local Lipschitz continuity in the system dynamics, which allows fitting a dynamics model while having explicit guarantees on where the learned model has bounded errors, namely in the neighborhood near the demonstration data. This insight enables corrective label generation with theoretical guarantees, potentially alleviating compounding errors in imitation learning.

% \begin{figure}[t]
%     \centering
%     % \begin{subfigure}{.48\linewidth}
%     %     \centering
%     %     \includegraphics[width=\linewidth]{media/lego.JPEG}
%     % \end{subfigure}
%     % \begin{subfigure}{.48\linewidth}
%     %     \centering
%     %     \includegraphics[width=\linewidth]{media/coin.JPEG}
%     % \end{subfigure}
%     % \caption{Two of the tasks we consider, \emph{GearInsertion} and \emph{GraspCoin}.}
%     \includegraphics[width=0.9\linewidth]{media/cover.jpg}
%     \caption{Our hardware platform and the various objects manipulated in our experiments, explained in detail in Fig.~\ref{fig:system}.}
%     \vspace{-.5em}
%     \label{fig:front}
% \end{figure}

Despite CCIL's effectiveness in simulation domains, there is an \textit{absence of empirical evidence} supporting its real-world applicability, and there remains a \textit{gap in understanding its parameter choices for practical application}. Critically, CCIL assumes the presence of local Lipschitz continuity~\cite{sastry2013nonlinear} in the system dynamics—a physical property that can vary across tasks and datasets. Many tasks, such as manipulation, involve complex contact interactions that exhibit discontinuities in dynamics, raising doubts about CCIL's efficacy under such challenging real-world conditions. Therefore, research questions persist about required design decisions, parameter choices, and tuning procedures to enable its success.  

Our work aims to answer these questions by comprehensively analyzing CCIL's performance on real-world fine manipulation tasks. Through extensive ablation studies, we explore CCIL's assumptions and examine the impact of design choices on its practicality. These studies give rise to our key insight, namely, that \textit{even in contact-rich manipulation domains with varying discontinuities in dynamical systems, it is feasible and practical to use CCIL to generate corrective labels in the sub-space that contains local continuity.} To this end, we propose a practical instantiation of the CCIL framework tailored to identify and fit dynamics models for domains with mixed local Lipschitz continuity. Our ablation study tests vary controlled variables---i.e., data quantity, dynamics model learning, and hyperparameter sensitivity---to produce practical guidelines for applying CCIL. Our experiments provide \textit{empirical evidence that generating corrective labels using the CCIL framework significantly benefits imitation learning}.

We focus on the domain of fine manipulation, an area that encompasses a wide range of real-world applications, from manufacturing to assistive robotics~\citep{sakurai2016thin, joseph2010chopstick,yamazaki2012autonomous}. Fine manipulation tasks are notably challenging  even for human experts due to their demand for precise control and susceptibility to failure from minor inaccuracies, resulting in limited data availability. However, learning from demonstration is often preferred over model-based methods for many fine manipulation tasks with hard-to-define cost functions (e.g., stitching wounds), or hard to model dynamics.  %This in turn constitutes a challenge to provide corrective labels manually even with interactive experts.
Thus, the inherent difficulties of fine manipulation underscore a critical need for strategies to mitigate compounding errors in imitation learning, making it an ideal testing ground.

In summary, we contribute:

\begin{itemize}[noitemsep, topsep=0pt,leftmargin=15pt]
    \item Experiments that demonstrate CCIL's capability to improve the performance of imitation learning agents on real-world fine manipulation tasks with limited data availability. CCIL boosts the success rate for GraspCube from 23\% to 83\%, for GearInsertion from 58\% to 72\%, and for GraspCoin from 17\% to 48\%.
    \item Experiments with varying availability of expert data that validate with statistical significance the performance boost CCIL offers in low-data regimes.
    %Our experiments  We provide empirical evidence that learning locally continuous dynamics models from limited data can improve imitation learning agents for contact-rich manipulation.
    \item Extensive ablation tests that analyze how design choices and hyperparameters affect the local Lipschitz continuity and label errors, key conditions for CCIL to succeed. Our insights provide practical guidance on choosing Lipschitz constraints for training dynamics and choosing error bounds to  generate corrective labels.  %Quantifying the performance gains from continuity-and model-based data augmentation across varying data quantity, environments, and hyperparameter choices.
    %\item Analyzing the key assumptions and practical considerations required to successfully apply CCIL to real-world robotic domains.
    %\item Offering insights into design choices like dynamics model learning, label generation, and filtering criteria that can maximize performance.
\end{itemize}

%% file: inputs/3-method.tex
\section{Continuity-based corrective labels}%\\for Imitation learning}

\begin{algorithm}[t]
\caption{Our Instantiation of CCIL: \textbf{C}ontinuity-based \textbf{C}orrective labels for \textbf{I}mitation \textbf{L}earning}
\label{alg:main}
\begin{algorithmic}[1]
    \STATE \textbf{Input:} $\mathcal{D^*}=(s^*_i, a^*_i, s^*_{i+1})$
    \STATE \textbf{Initialize:} $D^\mathcal{G} \leftarrow \varnothing$
    \STATE \texttt{// Learn Dynamics}
    \STATE $MSE\leftarrow\mathbb{E}_{(s^*_i,a^*_i,s^*_{i+1})\sim \mathcal{D}^*}\left[\hat{f}(s^*_i, a^*_i) + s^*_i - s^*_{i+1})\right]$
    \STATE $\hat{f}\leftarrow\arg\min_{\hat{f}} MSE \text{ s.t. } \|W\|_2\leq L$
    \STATE \texttt{// Generate Labels}
    \FOR{$i=1 .. n$}
      \STATE $(s^\mathcal{G}_i, a^\mathcal{G}_i) \leftarrow$ $\mathrm{GenLabels}$ $(s^*_i, a^*_i, s^*_{i+1})$
      \IF{$||J_{\hat{f}}(s^*_i, a^*_i)||_2\cdot||s^\mathcal{G}_i - s^*_i || < \epsilon$} \label{alg:main:label_rejection}
      \STATE $\mathcal{D^G} \leftarrow \mathcal{D^G} \cup (s^\mathcal{G}_i, a^\mathcal{G}_i)$
      \ENDIF
    \ENDFOR
    \STATE $\mathcal{D}\leftarrow\mathcal{D}^* \cup \mathcal{D}^\mathcal{G}$
    \STATE \texttt{// Learn Policy}
    \STATE $\mathrm{LearnPolicy}\ \pi$
    %\STATE \textbf{Output:} $D^\mathcal{G}$
  \hfill
 
    % \STATE \textbf{Function} $\mathrm{ LearnDynamics}$ $\hat{f}$
    % \STATE \quad Optimize a chosen objective from Sec.~\ref{sec:learn_dynamics}
    % \STATE \quad $MSE\leftarrow\mathbb{E}_{(s^*_i,a^*_i,s^*_{i+1})\sim \mathcal{D}^*}\left[\hat{f}(s^*_i, a^*_i) + s^*_i - s^*_{i+1})\right]$
    % \STATE \quad $\hat{f}\leftarrow\arg\min_{\hat{f}} MSE \text{ s.t. } \|W\|_2\leq L$
    % \STATE \textbf{Function} $\mathrm{GenDisturbedActionLabels}$ 
    % \STATE \quad $a^\mathcal{G}_i \leftarrow a^*_i + \Delta$, $\Delta \sim \mathcal{N}(0, \Sigma)$
    % \STATE \quad $s^\mathcal{G}_i \leftarrow \arg\min_{s^\mathcal{G}_i} ||s^\mathcal{G}_i+\hat{f}(s^\mathcal{G}_i, a^\mathcal{G}_i) - s^*_{i+1} ||$

    \STATE \textbf{Function} $\mathrm{GenLabels}$ 
    \STATE \quad $a^\mathcal{G}_i \leftarrow a^*_i$
    % \STATE \quad $s^\mathcal{G}_i \leftarrow \arg\min_{s^\mathcal{G}_i} ||s^\mathcal{G}_i+\hat{f}(s^\mathcal{G}_i, a^\mathcal{G}_i) -\textcolor{Highlight}{s^*_{i}} ||$
    \STATE \quad $s^\mathcal{G}_i \leftarrow s^*_i - \hat{f}(s^*_i,a^*_i)$

    \STATE \textbf{Function} $\mathrm{LearnPolicy}$
    \STATE \quad $L(a,\hat{a})\leftarrow$ policy loss function, as defined in Eq \ref{eq:policy_loss}
    \STATE \quad $\pi=\arg\min_{\pi}\mathbb{E}_{(s_i,a_i)\sim\mathcal{D}}\left[L(a_i, \pi(s_i))\right]$
\end{algorithmic}
\end{algorithm}

This section describes the framework we use to train  imitation learning policies and highlights the design choices we made. Our training framework uses behavior cloning \citep{alvinn} and Continuity-based Corrective Labels for Imitation Learning (CCIL). \textit{Behavior cloning} is an imitation learning algorithm that trains a policy from expert demonstrations, and \textit{CCIL} is a data augmentation framework to enhance the robustness of imitation learning agents. Our derivation follows the existing CCIL framework, but we provide a self-contained instantiation in Section~\ref{sub:ccil}. Choosing variants and parameters for this framework is largely domain-specific and can be costly to tune in the real world. We detail our practical instantiation in Section~\ref{sub:ccil_instance}. 

\subsection{Notation for Imitation Learning and Behavior Cloning}
We consider a finite-horizon Markov Decision Process (MDP), $\mathcal{M}=\left\{\mathcal{S}, \mathcal{A}, f, P_0\right\}$, where $\mathcal{S}$ is the state space, $\mathcal{A}$ is the action space, $f$ is the ground truth dynamics function,  and $P_0$ is the initial state distribution. A \textit{policy} maps a state to a distribution of actions $\pi: s \rightarrow a$. We assume a \emph{deterministic} world dynamics function, 
%\sandy{Verify edits.}
where $f$ maps a state $s_t$ and an action $a_t$ at time $t$ to the change in state such that the next state is  $s_{t+1} = s_t + f(s_t, a_t)$. Following the common setting in imitation learning, the true dynamics function $f$ is unknown, and we have only expert demonstrations $\mathcal{D^*}$ as a collection of transition triplets: $\mathcal{D^*}=\{(s^*_j,a^*_j,s^*_{j+1})\}_j$, where $s^*_{j+1} = s^*_j + f(s^*_j, a^*_j)$, i.e., for state $s^*_j$ the expert provided an action label of $a^*_j$. Behavior cloning learns a policy from these traces  by maximizing the likelihood of the expert actions being produced at the expert states:
$\arg \max_{\hat{\pi}} \mathbb{E}_{s^*_j, a^*_j, s^*_{j+1} \sim \mathcal{D}^*} \log (\hat{\pi}(a^*_j \mid s^*_j))$. 

\subsection{Continuity-based Corrective Labels}
\label{sub:ccil}

CCIL aims to reduce compounding errors in imitation learning by generating corrective labels that guide the robot back to expert-visited states and recovery from mistakes. For each transition in expert data, $(s^*_t, a^*_t, s^*_{t+1})$, CCIL proposes to generate some state-action pair $(s^\mathcal{G}_t,a^\mathcal{G}_t)$ to bring the agent to expert state $s^*_t$. With a known dynamics model $f$, we can sample labels and query $f$ to ensure $s^\mathcal{G}_t + f(s^\mathcal{G}_t,a^\mathcal{G}_t) - s^*_t \rightarrow 0$. Without it, CCIL must first train a dynamics model.

% \textbf{Training Dynamics Model} Following [%https://www.cs.cmu.edu/~arunvenk/papers/2016/Venkatraman_iser_16.pdf
% ], one can train a dynamics model by minimizing the following loss: 
\textbf{Training a Dynamics Model.} We can train a dynamics model by minimizing the following loss: 
\begin{equation}
    \label{eq:dynamics_objective}
     \mathbb{E}_{(s^*_t,a^*_t,s^*_{t+1})\sim \mathcal{D}^*}\left[\hat{f}(s^*_t, a^*_t) + s^*_t - s^*_{t+1}\right].
\end{equation}
A learned dynamics model can yield reliable predictions only near its data support, not on arbitrary states and actions. CCIL decides where to query the learned dynamics models by leveraging the presence of local Lipschitz continuity in the system dynamics, as specified in Definition \ref{def:local_lipschitz}. If $(s^*,a^*)$ belongs to a space with local Lipschitz continuity, given small changes in the input state or actions, the resultant changes in the system's state are bounded and predictable to a degree, as follows.
\begin{definition}[Local Lipschitz Continuity]
    \label{def:local_lipschitz}
    A function $f$ is \emph{locally Lipschitz continuous} around $(s,a)$ with coefficient $K$ if there exists a neighborhood of $(s,a)$ of size $\delta$ such that for every $(\tilde{s},\tilde{a})$ where $\|(s,a)-(\tilde{s},\tilde{a})\|\leq\delta$:
    \begin{equation}
        \|f(s,a)-f(\tilde{s},\tilde{a})\| \leq K\|(s,a)-(\tilde{s},\tilde{a})\|.
    \end{equation}
\end{definition}

% [wgan, guanya shi paper on drone flying]
CCIL encourages the learned dynamics function to exhibit local Lipschitz continuity by modifying the training objective. We choose spectral normalization regularization and elaborate on  the reasons in Sec.~\ref{sub:ccil_instance}. Specifically, to train a dynamics model $\hat{f}$ using a neural network of $n$-layers with weight matrices $W_1, W_2, \dots, W_n$, we can iteratively minimize the training objective (Eq.~\ref{eq:dynamics_objective}) while regularizing the model by setting
\begin{equation}
    W_i\leftarrow \frac{W_i}{\max(\|W_i\|_2, K^{-n})}\cdot K^{-n}
    \label{eq:spectral}
\end{equation}
for all $W_i$, where $K$ is the desired Lipschitz constant.

\textbf{Label Generation.} With a learned dynamics model $\hat{f}$, CCIL attempts to find a corrective label $(s^\mathcal{G}_t,a^\mathcal{G}_t)$ for each expert data point $(s^*_t,a^*_t)$ such that $s^\mathcal{G}_t + \hat{f}(s^\mathcal{G}_t,a^\mathcal{G}_t) \approx s^*_t$. We consider the BackTrack label generator inspired by Backward Euler used in modern simulators:
%  [cite mujoco simulator by emo todorov]
\begin{equation}
    \label{eq:backtrack_alg}
\begin{split}
    s^\mathcal{G}_t &\leftarrow s^*_t-\hat{f}(s^*_t,a^*_t) \\
    a^\mathcal{G}_t &\leftarrow a^*_t
\end{split}
\end{equation}

\textbf{Filtering Generated Labels by Label Distance.} The generated label ($s^\mathcal{G}_t, a^\mathcal{G}_t$) aims to bring the agent to some expert state $s^*_t$. Its error on the true dynamics is $\|f(s^\mathcal{G}_t,a^*_t)-\hat{f}(s^\mathcal{G}_t,a^*_t)\|$. CCIL provides a bound for this error, as follows.

\begin{theorem}
    \label{thm:label_quality_backtrack}
    When the dynamics model has a bounded training error $\epsilon$ on the training data, if the learned dynamics $\hat{f}$ and true dynamics $f$ are respectively locally $K_1$ and $K_2$-Lipschitz within some neighborhood of $(s^*_t,a^*_t)$ of size $\delta$, and $\|s^\mathcal{G}_t-s^*_t\|<\delta$, then
    \begin{equation}
        \label{eq:label_quality_backtrack}
        \left\|f\left(s_t^{\mathcal{G}}, a_t^*\right)-\hat{f}\left(s_t^{\mathcal{G}}, a_t^*\right)\right\| \leq \epsilon+\left(K_1+K_2\right)\left\|s_t^{\mathcal{G}}-s_t^*\right\|.
    \end{equation}
\end{theorem}

To ensure the generated labels are of low error bound, CCIL sets a hyperparameter $\delta$, denoting the desired size of the neighborhood for local Lipschitz continuity. It filters out generated labels outside the neighborhood $\|s^\mathcal{G}_t-s^*_t\| > \delta$. Sec.~\ref{sub:ccil_instance}
describes our modification to this filtering.
 
\textbf{Learn Policy}. A straightforward way to utilize the generated labels is to add them to the demonstration dataset to train any imitation learning agent. For simplicity, we consider the following behavior cloning objective to train a policy:
\begin{equation}
    \label{eq:bc_objective}
    \pi=\arg\min_{\pi}\mathbb{E}_{(s_i,a_i)\sim\mathcal{D}}\left[L(a_i, \pi(s_i))\right].
\end{equation}

\subsection{Our Practical Instantiation of CCIL}
\label{sub:ccil_instance}

CCIL's effectiveness critically hinges on (1) training a dynamics model for the locally continuous state-action space, and (2) using it to generate labels with reasonable error bounds. We recommend simple design choices based on practical insights. Alg. \ref{alg:main} summarizes our approach.

\textbf{Training Dynamic Models.} CCIL requires training a dynamics model that contains local Lipschitz continuity and proposes a few candidate loss functions. We aim to achieve this goal using only spectral normalization regularization. We highlight that the local Lipschitz constant $K$ varies not just between domains, but also between different state-actions in the same domain. Regardless of the training objective, as long as $\hat{f}$ is modeled via a neural network, we can measure the local Lipschitz constant at any state-action by calculating the local Jacobian matrix $J_{\hat{f}}$ as follows.
\begin{remark}
    \label{remark:local_L_jacobian}
    For a continuous function $\hat{f}$, the local Lipschitz constant in a neighborhood around a given point $(s,a)$ is
    \begin{equation}
        K(s,a) \approx \|J_{\hat{f}}(s,a)\|_2.
    \end{equation}
\end{remark}

Training with spectral normalization imposes an \emph{upper bound} on the Lipschitz constant. The learned model will still exhibit local Lipschitz continuity with varying Lipschitz constants at different state-actions. In practice, we can set a loose upper bound on the local Lipschitz continuity and still achieve local Lipschitz continuity at various regions in the state space, thereby satisfying the condition required by CCIL.

\textbf{Filtering Generated Labels by Label Error.} CCIL proposes to filter out generated labels $(s^\mathcal{G}_t, a^\mathcal{G}_t)$ if the label distance $\|s^\mathcal{G}_t-s^*_t\|$ violates a set threshold $ \|s^\mathcal{G}_t-s^*_t\| > \delta $. This implicitly assumes that the size of the neighborhood of local Lipschitz continuity stays constant throughout the state space, making it challenging to tune the threshold $\delta$. 
%\sandy{Should the following be added as a contribution in the Abstract and Intro?}
We refine this rejection process, as indicated in line \ref{alg:main:label_rejection} of Alg.~\ref{alg:main}. 

The error bound for each generated label (Theorem~\ref{thm:label_quality_backtrack}) depends not only on the label distance but also on the training error $\epsilon$ and the local Lipschitz coefficients of $\hat{f}$, $K_1$. By measuring $\epsilon$ and $K_1$ (Remark~\ref{remark:local_L_jacobian}), we can calculate an error bound for each generated label. For a synthetic label $(s^\mathcal{G}_t,a^\mathcal{G}_t)$ generated from expert demonstrations $(s^*_t,a^*_t)$, knowing $a^\mathcal{G}_t=a^*_t$ and assuming $\epsilon \approx 0$ and $K_2\propto K_1$, we can set a threshold on the acceptable label error bound $\sigma$ and reject the label if
\begin{equation}
    \|J_{\hat{f}}(s^*_t,a^*_t)\|_2\cdot \|s^\mathcal{G}_t - s^*_t\| \geq \sigma.
\end{equation}

In practice, we first generate all labels and examine the distribution of label error bounds. We then define $\sigma$ to be some quantile of the distribution. We effectively compute a ``trust'' region around the expert data to generate labels, and the size of the region varies for each data point. We visualize this region in empirical experiments in Fig.~\ref{fig:trust_region}. 

%% file: inputs/4-experiment.tex
\section{Experiment Design}

\begin{figure}[t!]
    \centering
    \vspace{.2em}
    \begin{subfigure}[t]{.36\linewidth}
        \centering
        \includegraphics[width=\linewidth]{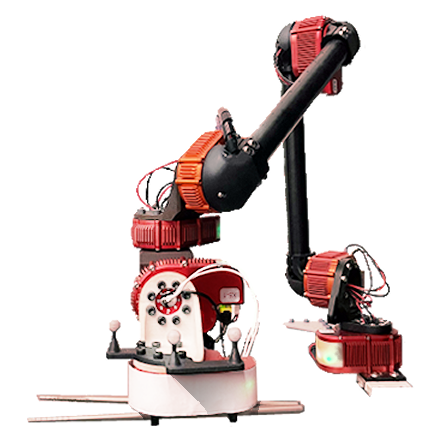}
         \caption{Robot Hardware}
         \label{fig:robot}
    \end{subfigure}
    % \begin{subfigure}[t]{.345\linewidth}
    %     \centering
    %     \includegraphics[width=\linewidth]{media/hardware_overview.jpg}
    %     \caption{System Overview}
    %     \label{fig:robot_mocap}
    % \end{subfigure}
    \begin{subfigure}[t]{.44\linewidth}
        \centering
        \includegraphics[width=.8\linewidth]{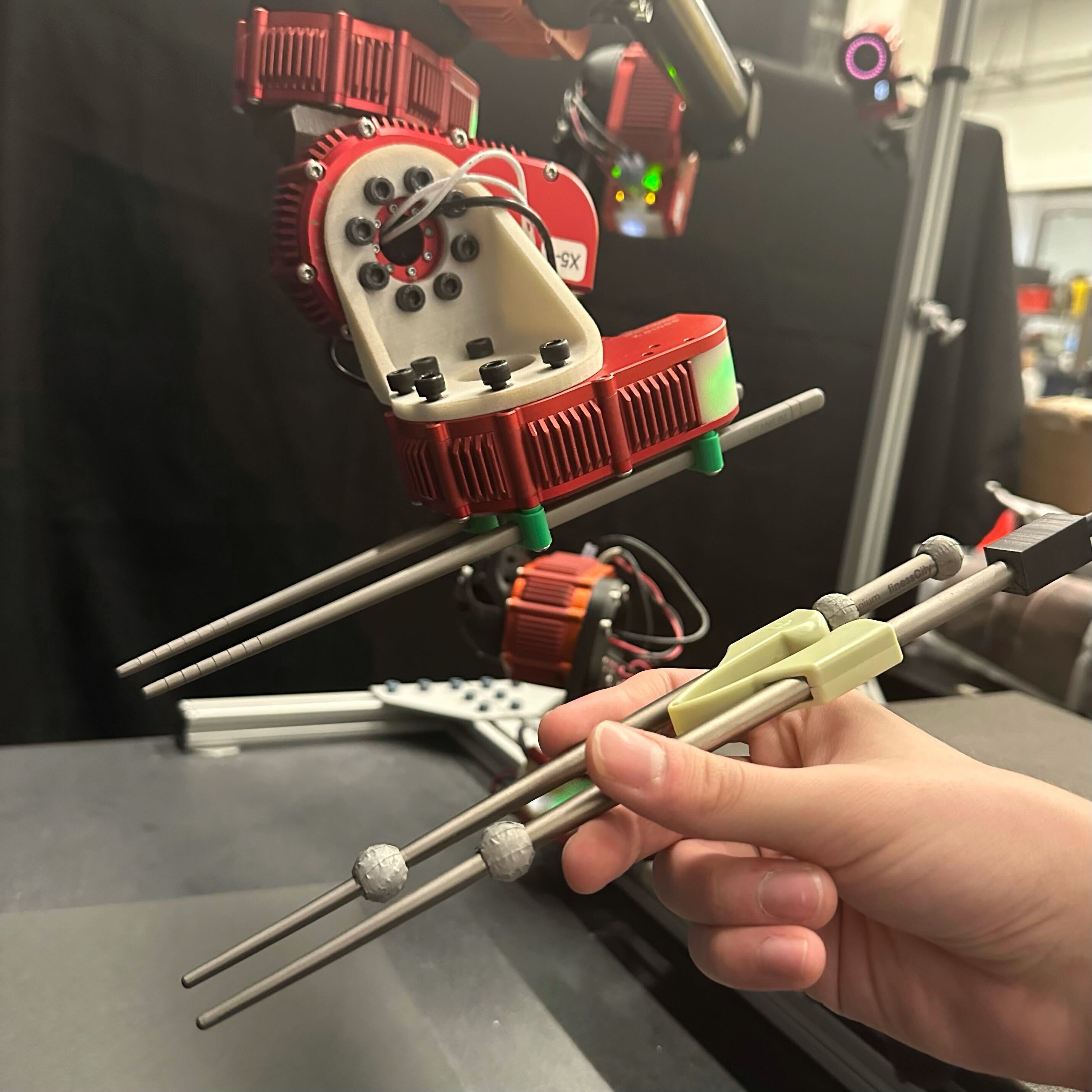}
        \caption{Teleoperation}
        \label{fig:teleop}
    \end{subfigure}
%     \begin{subfigure}[t]{.22\linewidth}
%         \centering
% \includegraphics[width=\linewidth]{media/objs_topdown_labeled.jpg}
%         \caption{Task Objects}
%         \label{fig:task_desc}
%     \end{subfigure}
    \caption{System overview. (a) Our HEBI-based 7-DOF robot with a chopstick end-effector. (b) Teleoperation mimicking leader chopsticks tracked using a motion-capture cage.}
    % \caption{System overview. In (d) we see the objects manipulated in our experiments. From left to right: a coin, 28mm in diameter and 3mm in height, whose thinness and rounded shape makes it difficult to grasp, even for humans. A gear and axle, 5mm in diameter, which must be inserted into a 5mm hole with great precision. A 1cm cube, whose flat edges make it easiest of the three to manipulate, although its small size still requires precise manipulation.}
    \vspace{-.5em}
    \label{fig:system}
\end{figure}

\subsection{Motivation}

The CCIL framework has shown notable success in various simulation domains. However, its application to real-world scenarios hinges on several critical research questions:

\textbf{Local Continuity in Dynamics and Real-World Application.} CCIL relies on local Lipschitz continuity in system dynamics, yet real-world robotic tasks often involve discontinuities due to physical contact. Can this foundational assumption be validated in realistic domains? Can CCIL still enhance imitation learning agent performance amidst such challenges?

\textbf{Data Scarcity and Augmentation Impact.} Limited real-world robot demonstrations highlight the importance of efficient data augmentation. While abundant data typically guarantees better outcomes, the critical question lies in the low-data regime: How effectively can CCIL address data scarcity, and what data volume is required to observe a tangible impact?

\textbf{Sensitivity to Hyperparameters.} Tuning imitation learning algorithms on physical robots is logistically challenging. Direct execution on robots, while being the most reliable evaluation method, risks unpredictable behaviors and necessitates numerous real-world trials to achieve statistical significance. How sensitive is CCIL to hyperparameter variations? Can we offer guidelines for its real-world application that mitigate the high evaluation and tuning costs?

\subsection{Hypotheses}

We explore the following hypotheses:
\begin{itemize}
    \item \textbf{H\hypothesis{hyp:data}:} CCIL can improve imitation learning policies in real-world fine-manipulation tasks. This improvement is statistically significant, especially in low-data regimes.

    \item \textbf{H\hypothesis{hyp:local_L}:} Real-world tasks with complex contacts and discontinuities in dynamics can still exhibit local Lipschitz continuity to justify CCIL's assumptions. %Consequently, CCIL can enhance the performance of imitation learning policies in such tasks.

    % \item \textbf{H\hypothesis{hyp:labels}:} Not all labels generated by CCIL are guaranteed to be useful. Incorporating low quality corrective labels can deteriorate the performance of imitation learning agents.
    % The performance of an imitation learning agent can improve as it incorporates more corrective labels. However, the performance will degrade if it incorporates low quality corrective labels that have large error bounds.

    \item \textbf{H\hypothesis{hyp:err_quantile}:} The performance of CCIL is highly sensitive to the label rejection hyperparameter, which determines the acceptable error bound for generated labels.

    \item \textbf{H\hypothesis{hyp:lipschitz}:} The performance of CCIL exhibits relative robustness to variations in the local Lipschitz constraint enforced during the training of the dynamics model.
    
    % A learned dynamics model effectively reflects a varying level of local Lipschitz continuity of the dynamics in different regions of the state-action space. Even with a relaxed Lipschitz constraint during training, the learned dynamics model will still exhibit local Lipschitz continuity. %, and the size of the continuity will converge.
    %Maybe, after we fix threshold of error bound, it does not matter how much Lipschitz continuity we enforce? We can use effective L and the fixed error bound to find the neighborhood?

    % \item \textbf{H\hypothesis{hyp:holdout}:} Adding corrective labels to the demonstration dataset does not inhibit a policy's ability to fit to the expert behavior.

\end{itemize}

To evaluate \hypref{hyp:data}, we compare the success rate of a behavior cloning agent trained with and without CCIL-generated labels. We vary the amount of demonstration data to investigate the impact of data quantity on these methods.

To evaluate \hypref{hyp:local_L}, we measure the local Lipschitz continuity of the trained dynamics model for tasks with complex discontinuity. We also vary the hyperparameter of the upper bound on the local Lipschitz constant. %By observing how continuous models fit to demonstration data, we can understand if our system dynamics are suitable for the application of CCIL.

% To support \hypref{hyp:labels}, we vary the number of corrective labels to include in training. By including synthetic labels with small or large label error, we test how CCIL reacts when given labels of varying quality.

% To investigate \hypref{hyp:lipschitz}, we ablate across several Lipschitz constraints and label error quantiles and measure the success rate of CCIL to understand the impact of these hyperparameters.
% we train the learned dynamics model with progressively weaker Lipschitz constraints. We can estimate the local Lipschitz coefficients at all data points, denoted as ``Effective'' local Lipschitz continuity. We observe how that distribution of effective local continuity varies with the strength of the Lipschitz constraint.

To evaluate \hypref{hyp:err_quantile} and \hypref{hyp:lipschitz}, we run ablation studies varying the hyperparameters of the label rejection threshold and the Lipschitz constraint to investigate (1) their impact on the success rate of CCIL, and (2) the interplay between these hyperparameters.

%To investigate \hypref{hyp:lipschitz}, we ablate across several Lipschitz constraints, and for each constraint value, test several label rejection quantils. This will show how Lipschitz constraints affect CCIL's ability to boost the performance of IL, and also how that constraint interplays with the label rejection quantile.

% To investigate \hypref{hyp:holdout}, we evaluate policies on a holdout dataset of expert demonstrations. By evaluating and comparing both a policy trained only on demonstrations against one trained with both demonstrations and corrective labels, we can understand how corrective labels impact a policy's ability to fit to expert demonstrations.

\subsection{Hardware}

We conducted experiments on a fine manipulator platform introduced in \cite{ke2021grasping} and shown in Fig.~\ref{fig:robot}. The robot has 7 joints and is equipped with a pair of chopsticks as its end-effector. It runs a hierarchical controller, where the high-level command consists of a 6-DOF end-effector pose target plus 1-DOF for the last joint to open or close the gripper chopsticks. A policy sends high-level commands at 20Hz. The robot computes inverse kinematics to send each joint a low-level positional command, which is tracked by a positional PID controller running at 1KHz. For perception, we use a motion capture system. We follow \cite{ke2020telemanipulation} to also collect human teleoperation demonstrations (Fig.~\ref{fig:teleop}).

\subsection{Tasks and Data Collection}
\label{sub:tasks}

We consider three fine manipulation tasks that require a millimeter level of precision. 
\begin{enumerate}[noitemsep, topsep=0pt,leftmargin=15pt]
    \item \emph{GraspCube}: Grasps and lifts a tiny 1cm cube above the table. Despite being the easiest of all three tasks, the task is non-trivial with scarce data (e.g., 100 trajectories). 
    \item \emph{GearInsertion}: Inserts a Lego gear into a hole on a board. This mini-gear insertion task (Fig.~\ref{fig:front}) is inspired by industrial assembly~\cite{gubbi2020imitation}. Proper insertion leaves a gap of less than 0.2 mm.  %When the insertion succeeds, the gear can rotate successfully to power the Lego windmill. 
    \item \emph{GraspCoin}: Uses chopsticks to grasp and lift a metal coin lying flat on the table. The coin's thin round shape and slippery texture makes it difficult even for human experts to pick up using chopsticks.  
\end{enumerate}

For all tasks, we vary the initial positions of the object. To collect demonstrations, we use a mix of heuristic controllers and human teleoperation. For \emph{GraspCube} and \emph{GearInsertion}, we designed heuristic controllers and collected 500 trajectories for \emph{GraspCube} and 100 trajectories for \emph{GearInsertion}. Note that the heuristic controllers did not have perfect success rates, and we filtered out failed demonstrations. For \emph{GraspCoin}, designing a heuristic controller was complicated due to the difficulty of the task, so we instead used 200 successful trajectories from an human expert teleoperation.

\subsection{Training}

For each task, we train two behavior cloning agents, with and without CCIL-generated labels. The training follows Alg.~\ref{alg:main}, and we formulate the action loss for our hardware and detail the parameter tuning procedure below.

% Data cleaning. Dynamics model training. How do we select hyper-parameters. What are intermediate plots we generate for selecting hyper-parameters. 

% Baselines: BC, NoiseBC?, CCIL.

%\textbf{Action Space}

%Let $Q=\{q\in \mathbb{R}^4\colon \|q\|=1\}$ be the set of unit quaternions. Our policies operate using position-control, and so our action space is the set of possible end-effector configurations. Concretely, $A=\mathbb{R}^3\times Q \times \mathbb{R}$. Note that this consists of a 6-DOF pose and the angle between the chopsticks.

\textbf{Loss Formulation.}
To train a policy using behavior cloning following Eq.~\ref{eq:bc_objective}, we need to design an action loss $L(a^*, \hat{a})$. We denote the action $a$ for our hardware as $(x,q,c)$, which represents the end-effector xyz location, orientation, and chopstick angle. Given the target action $a^*=(x,q,c)$ and the predicted action $\hat{a}=(\hat{x},\hat{q},\hat{c})$,  the action loss is:
\begin{equation}
    \label{eq:policy_loss}
    L(a,\hat{a})=\alpha_1\|x-\hat{x}\|^2 + \alpha_2\theta^2 + \alpha_3(c-\hat{c})^2,
\end{equation}
where $\theta$ is the angle between the orientations $q$ and $\hat{q}$ and $\alpha_1,\alpha_2,\alpha_3$ weight each component of the loss function. In our experiments, we use $\alpha_1=10,\alpha_2=1,\alpha_3=10$.

\textbf{Parameter Tuning}.  
%Since CCIL is primarily a data augmentation method, we use standard Behavior Cloning to train a policy. For a specific task, we use the same hyperparameters to train the policy between CCIL and BC, including the policy learning rate, batch size, policy architecture, and action loss hyperparameters.
Applying CCIL introduces two key parameters: (1) the Lipschitz constraint for training the dynamics model, and (2) a threshold for filtering the generated labels. The Lipschitz constraint, $K$ (Eq.~\ref{eq:spectral}), upper bounds the Lipschitz continuity of the learned dynamics model. A loose upper bound makes the training process easier and more likely to yield low training error. Conversely, a tighter bound is more likely to yield a dynamics model with lower Lipschitz coefficients, enabling CCIL to generate higher quality corrective labels. The label error bound is defined in Theorem.~\ref{thm:label_quality_backtrack}, and the threshold $\sigma$ controls the acceptable bound. %This value, together with the local Lipschitz coefficient at a given datapoint, constructs a trust region around that datapoint within which generated labels are accepted for training. 

%Concretely, when training a dynamics model, we first train an unconstrained dynamics model and plot the distribution of local Lipschitz coefficient across the training data, an example of which is shown in Fig.~\ref{fig:train_plots}. If this distribution has a significant tail, we then choose a Lipschitz constraint according to this plot that will eliminate any obvious outliers. After retraining with this constraint, we can regenerate the distribution plot and repeat until the distribution of local Lipschitz coefficients is uni-modal with a short tail. This parameter's behavior is further investigated in Section~\ref{sub:cont_constraint}.

%For filtering labels, we first generate corrective labels without any thresholding and generate a plot of the distribution of label error across the generated labels. Using this plot, we can choose a label rejection quantile that will filter out any groups of outliers or long tails in the distribution plot. We further explore this parameter in Section~\ref{sub:quality}.

To set the Lipschitz constraint $K$, we first train an unconstrained model and analyze the distribution of local Lipschitz coefficients on the learned model. % (Fig.\ref{fig:train_plots}). 
If this distribution suggests that the learned model has limited local Lipschitz continuity, we choose a tighter Lipschitz constraint.

To set threshold $\sigma$, we first generate corrective labels without filtering and analyze the distribution of label errors. Based on the distribution, we choose a label rejection quantile (between 0 to 1, where 0 means to filter out all generated labels) that filters out outliers or long tails. %Section~\ref{sub:quality} explores this parameter in depth.

\iffalse
\begin{figure}[h]
    \centering
        \includegraphics[width=0.45\linewidth]{figs/train_local_lipschitz.pdf}
        \includegraphics[width=0.45\linewidth]{figs/label_err_cdf.pdf}
    \caption{Example training plots generated for an unconstrained dynamics model. Right: The distribution of local Lipschitz coefficients across the demonstration dataset. Left: The CDF of the label error across the generated labels.}
    \label{fig:train_plots}
\end{figure}
\fi

\subsection{Evaluation}

We test each agent's success rate by conducting 48 trials per agent to establish statistical significance. To ensure fairness, for each task we select 16 fixed initial conditions (i.e., for grasping agents, we place the object to grasp at 16 fixed positions across the workspace). For each initial condition, we test the learned policy for 3 trials. For each trial, we denote the success as a binary variable. To report the statistical significance of the empirical results, we compare the success rates between the two policies by performing two-proportion z-tests.

%% file: inputs/5-results.tex
\section{Results}
\begin{figure*}[t!]
\vspace{.5em}
    \centering
    \begin{subfigure}[t]{.763\linewidth}
        \centering
        \includegraphics[width=0.327\linewidth]{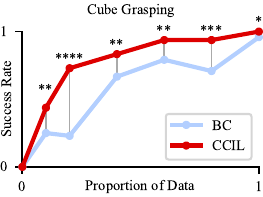}
        \includegraphics[width=0.327\linewidth]{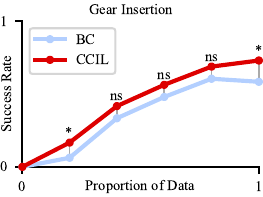}
        \includegraphics[width=0.327\linewidth]{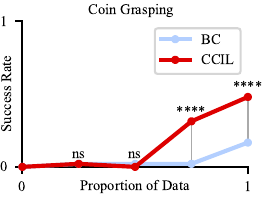}
        \caption{CCIL generally boosts performance over behavior cloning, especially in low-data regimes.}
        \label{fig:data_ablation}
    \end{subfigure}
    \begin{subfigure}[t]{.23\linewidth}
    \centering
    \includegraphics[width=\linewidth]{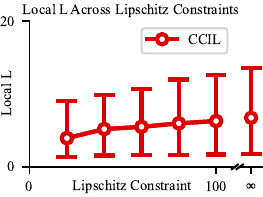}  
    \caption{CCIL is robust to Lipschitz.}
    \label{fig:local_L}\end{subfigure}
    \caption{(a) As the amount of data increases, CCIL provides a smaller boost over behavior cloning. We use asterisks to denote statistical significance: * p $< 0.1$, ** $p < 0.05$, *** $p < 0.01$, and **** $p < 0.001$. $ns$ denotes $p \geq 0.1$. (b) Mean and middle 95\% of the local Lipschitz coefficients of the learned dynamics model across the demonstration dataset as the Lipschitz constraint increases. As the enforced constraint increases, the distribution of coefficients converges.}
    \label{fig:ablation-and-lipschitz}
\end{figure*}

Across three tasks, we trained 36 agents and tested their success rate in the real world (Fig.~\ref{fig:data_ablation}). For ablation, we further evaluated 33 agents on the GraspCube task.  

\subsection{Corrective Labels' Improvement to Imitation Learning}

% \begin{figure*}[t!]
%     \centering
%     \includegraphics{figs/grasp_cube_ablation_line_plot.pdf}
%     \includegraphics{figs/lego_insert_ablation_line_plot.pdf}
%     \includegraphics{figs/coin_ablation_line_plot.pdf}
%     \caption{CCIL generally provides a performance boost over BC, especially in low-data regimes. As the amount of data increases, CCIL provides a smaller boost over BC. We use asterisks to denote statistical significance: * p $< 0.1$, ** $p < 0.05$, *** $p < 0.01$, and **** $p < 0.001$. $ns$ denotes $p \geq 0.1$.}
%     \vspace{-.5em}
%     \label{fig:data_ablation}
% \end{figure*}

\hypref{hyp:data} motivates the investigation of whether CCIL can improve imitation learning performance in real-world fine-manipulation tasks. These tasks involve complex contact dynamics between objects, end-effectors, and the workspace, making it unclear if CCIL's assumptions hold.

%We train and evaluate CCIL and BC on three challenges and report the success rates for each agent in Fig.~\ref{}. Furthermore, to understand the impact of data scarcity on CCIL's performance, we ablate on the number of expert demonstrations to use, and report the performance in Fig.~\ref{fig:data_ablation}.

\textbf{CCIL yields an increase in performance for behavior cloning.} Fig.~\ref{fig:data_ablation} shows that training with labels generated by CCIL generally improves the performance of behavior cloning over all three tasks, as measured by success rate. We validated the statistical significant of the improvement. The performance boost is significant at the $p<0.05$ level for the cube grasping and coin grasping tasks.

% On three tasks, training with CCIL generated labels generally improves the performance of behavior cloning, judged by success rate. We validated that the improvement is statistically significant on two of the tasks (XY) with 95. For the third task, ..... Our experiments testified the real-world applicability of CCIL to complex fine manipulation tasks that contain discontinuity. 

% \textbf{CCIL proved to be effective in low-data regime} We ablate the quantity of data for imitation learning training. 
% Training on varying fractions of the dataset for each task, we see that CCIL consistently at least matches the performance of BC. We can attribute this partially to the fact that for low values of a label error rejection quantile, CCIL produces a small number of extremely confident labels in the nearby vicinity of demonstration data, and so converges to the performance of BC.

\textbf{The CCIL boost is significant in the low-data regime}. The empirical performance gain from applying CCIL is statistically significant in low-data regimes, as shown in Fig.~\ref{fig:data_ablation}. CCIL boosts behavior cloning performance from 23\% to 83\% (using 100 GraspCube trajectories), from 6\% to 17\% (using 20 GearInsertion trajectories) or 58\% to 72\% (using 100 GearInsertion trajectories), and from 17\% to 48\% (using 200 GraspCoin trajectories). These limited-data regimes face substantial challenges from covariate shift due to limited data support from expert demonstrations. CCIL demonstrated promising results to alleviate this problem.   %With limited data, covariate shift is a significant challenge in this regime, as there are not enough expert demonstrations to provide coverage over the state distribution visited by the policy. CCIL helps to alleviate this problem, since the generated labels serve to synthetically expand the support of the demonstration data. When trained with this increased coverage, policies are able to reduce covariate shift and perform these tasks with greater success.

\textbf{Remark}. Our experiments validate how CCIL can be applied to complex fine manipulation tasks in the real world despite the presence of discontinuity in the contact dynamics.

% Additionally, it is important to note that CCIL does not make guarantees about the effect of the corrective labels on policy performance. Therefore, as seen in the lego insertion task, there is no guarantee that these corrective labels will boost performance. We further investigate these phenomena in \ref{}.

% There is a substantial performance (Fig). This highlight the benefit of using XXX to boost YYY. The performance gap is large at ... Qualitatively what did you observe? Interestingly, using human data, boost is larger (replace with your finding)

% Offline data driven methods are hindered by scarcity of data and limited data support. Interestingly CCIL was able to boost in low data region. Talk about the chart, the numbers, The statistical significance. Surprisingly,... talk about your finding.

\subsection{CCIL's Assumptions on Local Lipschitz Continuity}

\hypref{hyp:local_L} motivates the investigation of  whether CCIL's local continuity assumptions are sufficiently valid and realistic to apply to real-world fine manipulation tasks. 

\begin{figure*}
    \centering
    \begin{subfigure}[t]{0.3\textwidth}
        \centering
        \includegraphics[width=\linewidth]{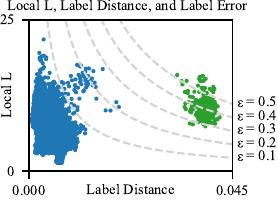}
        \caption{Clusters reveal dynamics regimes.}
        \label{fig:error_scatter}
    \end{subfigure}
    \hfill{}
    \begin{subfigure}[t]{0.3\textwidth}
        \centering
        \includegraphics[width=\linewidth]{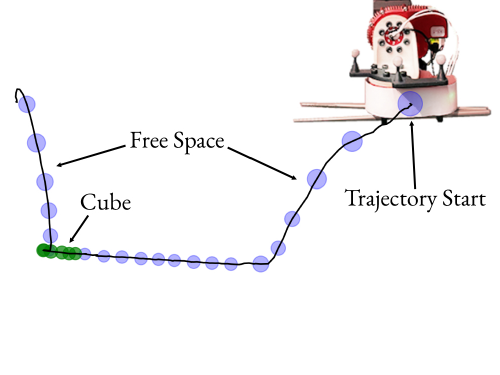}
        \caption{High label rejection clusters around contact.}
        \label{fig:trust_region}
    \end{subfigure}
    \hfill{}
    \begin{subfigure}[t]{0.3\textwidth}
        \centering
        \includegraphics[width=\linewidth]{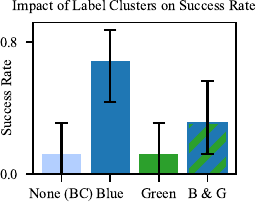}
        \caption{}
        \label{fig:cluster_ablation}
    \end{subfigure}
    \caption{(a) The 20\% dataset for the \emph{GraspCube} task reveals two distinct clusters of corrective labels. (b) The green cluster mainly  corresponds to labels where the cube is being manipulated, and the blue cluster to arm-free space motion. (c) Policies trained using just the blue cluster (low label rejection threshold) are more successful compared to  those trained with the green cluster or both (high label rejection threshold).}
\end{figure*}

\textbf{Real-world tasks contain local Lipschitz continuity that can be realized by learning dynamics models.} 
In training the dynamics model, we experiment with different Lipschitz constraints, measure the resulting models' local Lipschitz coefficient, and plot the average of the coefficient in Fig.~\ref{fig:local_L}. With a large upper bound on the Lipschitz constraint (loose constraint), the local coefficients increase but converge to those of an unconstrained model. These findings imply that our environments exhibit local continuity that the learned dynamics models are fitting to, even without explicitly enforcing the Lipschitz continuity coefficient. 

\textbf{The learned dynamics model and generated labels have varying continuity that correlates with the real world.} For each generated label, we can measure the local Lipschitz coefficients using the learned dynamics model (``Local L''). We generate a scatterplot for the generated labels in Fig.~\ref{fig:error_scatter} and see two clusters (green and blue). We then sample some of the labels for visualization and plot them along an expert trajectory in Fig.~\ref{fig:trust_region}. We observe that the green cluster corresponds  to labels generated near the cube, and the blue labels are mainly in free space when the robot moves without collision. Intuitively, the learned model and generated label have higher errors for the green group that contains discontinuity.

We test how policy performance changes as we incorporate into the demonstration dataset (1) blue labels, (2) green labels, and (3) both labels, as shown in Fig.~\ref{fig:cluster_ablation}. Training with blue labels improves the agent's success rate. However, training with the green labels associated with contact-rich states and higher errors is not as useful. Such labels actually diminish performance. This highlights how label quality can impact CCIL performance and the significance of filtering generated labels.
%\sandy{Do you want a Remark summary here as in prior section?}

\subsection{Impact of the Quality of Generated Labels}
\label{sub:quality}

\begin{figure}[t!]
    \centering
    \includegraphics{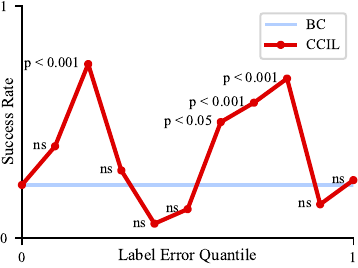}
    \caption{Policy performance when filtering out varying fractions of the generated labels due to label error.}
    \vspace{-.5em}
    \label{fig:quantile_ablation}
\end{figure}

The label filtering threshold controls what generated labels are used when training a policy. It is therefore crucial to understand its effects and how to tune it properly. We propose a practical way to enforce a filtering threshold via label rejection quantile; \hypref{hyp:err_quantile} motivated exploring how this design choice affects CCIL's performance, which we do below.

We choose the GraspCube task with 100 trajectories. We evaluate running CCIL with different label rejection quantiles and report the success rate in Fig.~\ref{fig:quantile_ablation}. The rejection quantile controls how many generated labels are used (25\%, 75\%, etc.) based on their computed error bound.  

%\textbf{When the label rejection quantile is near-zero, CCIL performs similarly to BC}. As the quantile goes to zero, the policy performance converges to that of BC. This is because, at 0, no generated labels are used. As the label error quantile decreases, we reject more and more labels. So when no labels are used for policy training, CCIL reduces exactly to BC.

\textbf{Low-quality labels can harm policy performance.} Using all generated labels (i.e., choosing a quantile close or equal to 1) makes CCIL performance similar to or worse than behavior cloning. With this quantile, the generated labels have high error bounds and could be incorrect under the true dynamics. If these labels are included when training the policy, they could potentially diminish policy performance.

\textbf{There is a need to balance label error and usefulness for CCIL to succeed}. Labels with lower error bounds are more trustworthy. However, these labels tend to be closer to the demonstration data (small label distance). Including labels with conservative error bounds may fail to expand the data support. Conversely, a label that is further from the demonstration data can expand the data support while potentially introducing a higher associated error. Fig.~\ref{fig:quantile_ablation} shows intermediate values of the label error quantile that achieve a higher success rate, indicating the need to balance label accuracy and usefulness to maximize CCIL's performance.
%\sandy{Do you want a Remark summary here?}

\subsection{Impact of Continuity Constraint}
\label{sub:cont_constraint}

\begin{figure}[H]
    \centering
    \includegraphics[width=\columnwidth]{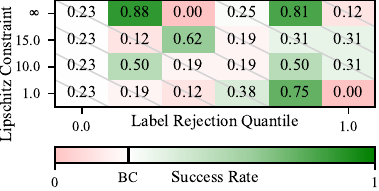}
    \caption{Hyperparameter ablation for CCIL, varying the Lipschitz constraint and the label rejection quantile. Green cells indicate a performance boost over behavior cloning, while red cells show worse performance. Crossed out cells are not significant at the $p<0.05$ level, while the others are.}
    \label{fig:hparam_ablation}
\end{figure}

\hypref{hyp:lipschitz} asks how enforcing Lipschitz continuity for training the dynamics model impacts CCIL's performance. %Continuity can vary not only between environments, but between different tasks in the same environment. Thus, understanding this hyperparameter's impact is important to apply CCIL to other domains.

We train several dynamics models using different Lipschitz constraints and test CCIL's performance while varying the label rejection quantiles. %As before, we fix the task to cube grasping and the dataset to 20\%. 
%All hyperparameters other than those ablated upon are fixed. 
We show the resulting success rate in Fig.~\ref{fig:hparam_ablation}, crossing out the cells whose performance is not statistically significant compared to behavior cloning. 

% \textbf{Tuning the label error quantile is crucial for performance.} Looking across each row of the grid in Fig.~\ref{fig:hparam_ablation}, we see that policy performance can vary greatly with label rejection quantile. As discussed in subsection \ref{sub:quality}, performance at 0.0 is the same as BC, and performance at 1.0 can potentially be worse. In between, though, we see that tuning the label error quantile appropriately can result in great improvements over BC.

\textbf{CCIL can work with different Lipschitz constraints.} For each Lipschitz constraint, we observe a label rejection quantile that achieves a significant boost in performance over behavior cloning. This indicates that CCIL could be less sensitive to the enforced Lipschitz continuity parameter.

\textbf{CCIL can work with unconstrained dynamics models.} Surprisingly, even with an unconstrained dynamics model, as indicated by a Lipschitz constraint of $\infty$, CCIL still performs well at 20\% and 80\% quantile. We investigate this phenomenon in Fig.~\ref{fig:local_L} %by ablating across different Lipschitz constraints and plotting the average local Lipschitz coefficient of the learned dynamics model across the demonstration data, in addition to the middle 95\% of local Lipschitz coefficients. 
and observe that the unconstrained model still exhibits some local continuity, as shown by the relatively small local Lipschitz coefficients across the dataset. We see that as the Lipschitz constraint relaxes, the distribution of local Lipschitz coefficients converges to that of the unconstrained model.

\textbf{Remark}. These findings together imply that in environments with sufficient local continuity, explicit Lipschitz constraint enforcement may be unnecessary and tuning the label rejection quantile can be more critical. When applying CCIL to new environments, we can start with  unconstrained dynamics models for their simplicity. %unconstrained models should be preferred initially, due to their simplicity and lack of Lipschitz constraint to tune.

%% file: inputs/6-discussion.tex
\section{Conclusion}

This paper demonstrated the effectiveness of corrective labels for offline imitation learning through real-world fine-manipulation experiments, highlighting its impact in the low-data regime. Our results validate that the local continuity assumptions are applicable even in the presence of discontinuous contact dynamics. Furthermore, we provide practical guidelines for choosing the continuity constraint and label quality threshold to enable CCIL's success. Future work can explore extending CCIL to high-dimensional state spaces, such as images, and investigating different policy classes, such as  diffusion policies.
 %interact with each other and impact CCIL's performance, which is crucial to understand in order to apply CCIL effectively.

%We observe that CCIL is a practical and intuitive data augmentation method with minimal assumptions, and we believe its adoption could greatly aid in combating data scarcity and covariate shift in future work in offline Imitation Learning.

\iffalse

\subsection{Limitations}

CCIL is not without its drawbacks, however. Due to its minimal additional assumptions, CCIL does not provide any proven bounds on how its corrective labels will impact a policy's performance. Although these labels boost performance empirically, our experiments show that it is nontrivial to predict how different labels will impact a policy.

\subsection{Future Work}

This work, to our knowledge, marks the first deep investigation of corrective labels for Imitation Learning on real-world platforms. However, there still remain many new and exciting avenues to further explore this problem setting.
\begin{itemize}
    \item Our experiments use continuity in low-dimensional state to generate corrective labels. If future work can apply CCIL to complex and high-dimensional inputs such as images, it could greatly improve CCIL's applicability.
    \item CCIL is one of multiple data augmentation methods for Imitation Learning, and it would be fruitful to explore how these methods can work together to improve policy performance.
\end{itemize}
\fi

%% file: inputs/2-related-work.tex
\section{Related Work}
\textbf{Imitation Learning and Behavior Cloning.}
\textit{Imitation learning} can enable robots to mimic complex behaviors from expert demonstrations~\citep{billard2008survey, osa2018algorithmic}. \textit{Behavior cloning }(BC)~\citep{alvinn} is a simple and practical imitation learning method that requires only demonstration data to learn a robotics policy, remaining a strong candidate for practical application and therefore gaining popularity over the years~\citep{bojarski2016end, zhang2018deep, kalashnikov2018scalable, rhinehart2018deep, chi2023diffusion}. Empirical studies have reported performance gaps between simulation and real world results and highlighted the importance of real-world evaluation~\citep{mandlekar2021matters, zhou2023real}. 

\textbf{Covariate Shift and Data Augmentation.} 
Compounding errors in robot execution can 
cause agents to encounter unexpected and out-of-distribution states~\citep{osa2018algorithmic}. Various works that address covariate shift often require additional information, including data from interactive experts~\citep{ross2011dagger,laskey2017dart,sun2017aggrevated}, environmental samples~\citep{ho2016gail,il_f_divergence, fu2017airl}, alternative sources~\citep{chang2021mitigating} and domain knowledge~\citep{ke2021grasping, zhou2023nerf, chen2023genaug, block2024provable, ke2023ccil}. To achieve broad applicability without such prerequisites, we focus on CCIL, an imitation learning method that operates under minimal assumptions, relying solely on offline demonstrations without requiring interactive experts or simulators. Instead, it makes a light assumption about the presence of local Lipschitz continuity. Despite promising results, there is scant empirical evidence for CCIL's real-world applicability, which we aim to provide.% We aim to provide this evidence.

\textbf{Local Lipschitz Continuity in Dynamics}. Classical control methods and modern robot applications~\citep{bonnard2007second, li2004iterative, kahveci2007robust} often assume local Lipschitz continuity in system dynamics to guarantee solution existence and uniqueness. However, they operate with pre-specified models and cost functions. Most works on learning dynamics from data neither leverage nor enforce the local Lipschitz continuity constraint~\citep{hafner2019dream, wang2022denoised, kaufmann2023champion}. In contrast, this work focuses on learning locally continuous dynamics models from data without requiring human-specified models. While some prior works have enforced Lipschitz constraints during modeling of training dynamics~\citep{shi2019neural, pfrommer2021contactnets}, their application focuses on global continuity assumptions or simulation scenarios. It remains unclear how to learn dynamics models that capture the complex contacts and high precision requirements of real-world fine manipulation tasks while achieving local Lipschitz continuity, a key focus of our exploration in this paper.

%% file: inputs/999-acknowledgement.tex
\section{Acknowledgement}

This work was (partially) funded by the National Science Foundation NRI (\#2132848) and CHS (\#2007011), DARPA RACER (\#HR0011-21-C-0171), the Office of Naval Research (\#N00014-17-1-2617-P00004 and \#2022-016-01 UW), Amazon and funding from the Toyota Research Institute through the University Research Program 3.0. 

%% file: main.bbl
\begin{thebibliography}{10}
\providecommand{\url}[1]{#1}
\csname url@samestyle\endcsname
\providecommand{\newblock}{\relax}
\providecommand{\bibinfo}[2]{#2}
\providecommand{\BIBentrySTDinterwordspacing}{\spaceskip=0pt\relax}
\providecommand{\BIBentryALTinterwordstretchfactor}{4}
\providecommand{\BIBentryALTinterwordspacing}{\spaceskip=\fontdimen2\font plus
\BIBentryALTinterwordstretchfactor\fontdimen3\font minus \fontdimen4\font\relax}
\providecommand{\BIBforeignlanguage}[2]{{%
\expandafter\ifx\csname l@#1\endcsname\relax
\typeout{** WARNING: IEEEtran.bst: No hyphenation pattern has been}%
\typeout{** loaded for the language `#1'. Using the pattern for}%
\typeout{** the default language instead.}%
\else
\language=\csname l@#1\endcsname
\fi
#2}}
\providecommand{\BIBdecl}{\relax}
\BIBdecl

\bibitem{alvinn}
D.~A. Pomerleau, ``Alvinn: An autonomous land vehicle in a neural network,'' 1988.

\bibitem{floridi2020gpt}
L.~Floridi and M.~Chiriatti, ``Gpt-3: Its nature, scope, limits, and consequences,'' \emph{Minds and Machines}, 2020.

\bibitem{padalkar2023open}
A.~Padalkar, A.~Pooley, A.~Jain, A.~Bewley, A.~Herzog, A.~Irpan, A.~Khazatsky, A.~Rai, A.~Singh, A.~Brohan \emph{et~al.}, ``Open x-embodiment: Robotic learning datasets and rt-x models,'' 2023.

\bibitem{chi2023diffusion}
C.~Chi, S.~Feng, Y.~Du, Z.~Xu, E.~Cousineau, B.~Burchfiel, and S.~Song, ``Diffusion policy: Visuomotor policy learning via action diffusion,'' 2023.

\bibitem{billard2012imitation}
A.~Billard and D.~Grollman, ``Imitation learning in robots,'' \emph{Encyclopedia of the Sciences of Learning}, 2012.

\bibitem{osa2018algorithmic}
T.~Osa, J.~Pajarinen, G.~Neumann, J.~A. Bagnell, P.~Abbeel, J.~Peters \emph{et~al.}, ``An algorithmic perspective on imitation learning,'' \emph{Foundations and Trends{\textregistered} in Robotics}, 2018.

\bibitem{spencer2021feedback}
J.~Spencer, S.~Choudhury, A.~Venkatraman, B.~Ziebart, and J.~A. Bagnell, ``Feedback in imitation learning: The three regimes of covariate shift,'' 2021.

\bibitem{ross2011reduction}
S.~Ross, G.~Gordon, and D.~Bagnell, ``A reduction of imitation learning and structured prediction to no-regret online learning.''\hskip 1em plus 0.5em minus 0.4em\relax JMLR, 2011.

\bibitem{bryan2019data}
N.~J. Bryan, ``Data augmentation and deep convolutional neural networks for blind room acoustic parameter estimation,'' 2019.

\bibitem{wang2024cyberdemo}
J.~Wang, Y.~Qin, K.~Kuang, Y.~Korkmaz, A.~Gurumoorthy, H.~Su, and X.~Wang, ``Cyberdemo: Augmenting simulated human demonstration for real-world dexterous manipulation,'' 2024.

\bibitem{chang2021mitigating}
J.~Chang, M.~Uehara, D.~Sreenivas, R.~Kidambi, and W.~Sun, ``Mitigating covariate shift in imitation learning via offline data with partial coverage,'' 2021.

\bibitem{venkatraman2014data}
A.~Venkatraman, B.~Boots, M.~Hebert, and J.~A. Bagnell, ``Data as demonstrator with applications to system identification,'' in \emph{ALR Workshop, NIPS}, 2014.

\bibitem{ke2021grasping}
L.~Ke, J.~Wang, T.~Bhattacharjee, B.~Boots, and S.~Srinivasa, ``Grasping with chopsticks: Combating covariate shift in model-free imitation learning for fine manipulation,'' 2021.

\bibitem{zhou2023nerf}
A.~Zhou, M.~J. Kim, L.~Wang, P.~Florence, and C.~Finn, ``Nerf in the palm of your hand: Corrective augmentation for robotics via novel-view synthesis,'' 2023.

\bibitem{chen2023genaug}
Z.~Chen, S.~Kiami, A.~Gupta, and V.~Kumar, ``Genaug: Retargeting behaviors to unseen situations via generative augmentation,'' 2023.

\bibitem{block2024provable}
A.~Block, A.~Jadbabaie, D.~Pfrommer, M.~Simchowitz, and R.~Tedrake, ``Provable guarantees for generative behavior cloning: Bridging low-level stability and high-level behavior,'' 2024.

\bibitem{ke2023ccil}
L.~Ke, Y.~Zhang, A.~Deshpande, S.~Srinivasa, and A.~Gupta, ``Ccil: Continuity-based data augmentation for corrective imitation learning,'' 2023.

\bibitem{sastry2013nonlinear}
S.~Sastry, \emph{Nonlinear systems: analysis, stability, and control}.\hskip 1em plus 0.5em minus 0.4em\relax Springer Science \& Business Media, 2013, vol.~10.

\bibitem{sakurai2016thin}
H.~Sakurai, T.~Kanno, and K.~Kawashima, ``Thin-diameter chopsticks robot for laparoscopic surgery.''\hskip 1em plus 0.5em minus 0.4em\relax IEEE, 2016.

\bibitem{joseph2010chopstick}
R.~A. Joseph, A.~C. Goh, S.~P. Cuevas, M.~A. Donovan, M.~G. Kauffman, N.~A. Salas, B.~Miles, B.~L. Bass, and B.~J. Dunkin, ``“chopstick” surgery: a novel technique improves surgeon performance and eliminates arm collision in robotic single-incision laparoscopic surgery,'' 2010.

\bibitem{yamazaki2012autonomous}
A.~Yamazaki and R.~Masuda, ``Autonomous foods handling by chopsticks for meal assistant robot.''\hskip 1em plus 0.5em minus 0.4em\relax VDE, 2012.

\bibitem{ke2020telemanipulation}
L.~Ke, A.~Kamat, J.~Wang, T.~Bhattacharjee, C.~Mavrogiannis, and S.~S. Srinivasa, ``Telemanipulation with chopsticks: Analyzing human factors in user demonstrations.''\hskip 1em plus 0.5em minus 0.4em\relax IEEE, 2020.

\bibitem{gubbi2020imitation}
S.~Gubbi, S.~Kolathaya, and B.~Amrutur, ``Imitation learning for high precision peg-in-hole tasks,'' in \emph{ICCAR}, 2020.

\bibitem{billard2008survey}
A.~Billard, S.~Calinon, R.~Dillmann, and S.~Schaal, ``Survey: Robot programming by demonstration,'' 2008.

\bibitem{bojarski2016end}
M.~Bojarski, D.~Del~Testa, D.~Dworakowski, B.~Firner, B.~Flepp, P.~Goyal, L.~D. Jackel, M.~Monfort, U.~Muller, J.~Zhang \emph{et~al.}, ``End to end learning for self-driving cars,'' 2016.

\bibitem{zhang2018deep}
T.~Zhang, Z.~McCarthy, O.~Jow, D.~Lee, X.~Chen, K.~Goldberg, and P.~Abbeel, ``Deep imitation learning for complex manipulation tasks from virtual reality teleoperation,'' 2018.

\bibitem{kalashnikov2018scalable}
D.~Kalashnikov, A.~Irpan, P.~Pastor, J.~Ibarz, A.~Herzog, E.~Jang, D.~Quillen, E.~Holly, M.~Kalakrishnan, V.~Vanhoucke \emph{et~al.}, ``Scalable deep reinforcement learning for vision-based robotic manipulation.''\hskip 1em plus 0.5em minus 0.4em\relax PMLR, 2018.

\bibitem{rhinehart2018deep}
N.~Rhinehart, R.~McAllister, and S.~Levine, ``Deep imitative models for flexible inference, planning, and control,'' 2018.

\bibitem{mandlekar2021matters}
A.~Mandlekar, D.~Xu, J.~Wong, S.~Nasiriany, C.~Wang, R.~Kulkarni, L.~Fei-Fei, S.~Savarese, Y.~Zhu, and R.~Mart{\'\i}n-Mart{\'\i}n, ``What matters in learning from offline human demonstrations for robot manipulation,'' 2021.

\bibitem{zhou2023real}
G.~Zhou, L.~Ke, S.~Srinivasa, A.~Gupta, A.~Rajeswaran, and V.~Kumar, ``Real world offline reinforcement learning with realistic data source,'' in \emph{ICRA}, 2023.

\bibitem{ross2011dagger}
S.~Ross, G.~Gordon, and D.~Bagnell, ``A reduction of imitation learning and structured prediction to no-regret online learning.''\hskip 1em plus 0.5em minus 0.4em\relax JMLR, 2011.

\bibitem{laskey2017dart}
M.~Laskey, J.~Lee, R.~Fox, A.~Dragan, and K.~Goldberg, ``Dart: Noise injection for robust imitation learning.''\hskip 1em plus 0.5em minus 0.4em\relax PMLR, 2017.

\bibitem{sun2017aggrevated}
W.~Sun, A.~Venkatraman, G.~J. Gordon, B.~Boots, and J.~A. Bagnell, ``Deeply aggrevated: Differentiable imitation learning for sequential prediction.''\hskip 1em plus 0.5em minus 0.4em\relax PMLR, 2017.

\bibitem{ho2016gail}
J.~Ho and S.~Ermon, ``Generative adversarial imitation learning,'' \emph{NeurIPS}, 2016.

\bibitem{il_f_divergence}
L.~Ke, S.~Choudhury, M.~Barnes, W.~Sun, G.~Lee, and S.~Srinivasa, ``Imitation learning as f-divergence minimization.''\hskip 1em plus 0.5em minus 0.4em\relax Springer, 2021.

\bibitem{fu2017airl}
J.~Fu, K.~Luo, and S.~Levine, ``Learning robust rewards with adversarial inverse reinforcement learning,'' 2017.

\bibitem{bonnard2007second}
B.~Bonnard, J.-B. Caillau, and E.~Tr{\'e}lat, ``Second order optimality conditions in the smooth case and applications in optimal control,'' \emph{ESAIM: Control, Optimisation and Calculus of Variations}, 2007.

\bibitem{li2004iterative}
W.~Li and E.~Todorov, ``Iterative linear quadratic regulator design for nonlinear biological movement systems.'' in \emph{ICINCO (1)}, 2004.

\bibitem{kahveci2007robust}
N.~E. Kahveci, ``Robust adaptive control for unmanned aerial vehicles,'' Ph.D. dissertation, 2007.

\bibitem{hafner2019dream}
D.~Hafner, T.~Lillicrap, J.~Ba, and M.~Norouzi, ``Dream to control: Learning behaviors by latent imagination,'' 2019.

\bibitem{wang2022denoised}
T.~Wang, S.~S. Du, A.~Torralba, P.~Isola, A.~Zhang, and Y.~Tian, ``Denoised mdps: Learning world models better than the world itself,'' 2022.

\bibitem{kaufmann2023champion}
E.~Kaufmann, L.~Bauersfeld, A.~Loquercio, M.~M{\"u}ller, V.~Koltun, and D.~Scaramuzza, ``Champion-level drone racing using deep reinforcement learning,'' \emph{Nature}, 2023.

\bibitem{shi2019neural}
G.~Shi, X.~Shi, M.~O’Connell, R.~Yu, K.~Azizzadenesheli, A.~Anandkumar, Y.~Yue, and S.-J. Chung, ``Neural lander: Stable drone landing control using learned dynamics,'' 2019.

\bibitem{pfrommer2021contactnets}
S.~Pfrommer, M.~Halm, and M.~Posa, ``Contactnets: Learning discontinuous contact dynamics with smooth, implicit representations.''\hskip 1em plus 0.5em minus 0.4em\relax PMLR, 2021.

\end{thebibliography}
